# COMPARISON OF DIFFERENT METHODS FOR TISSUE SEGMENTATION IN HISTOPATHOLOGICAL WHOLE-SLIDE IMAGES

*Péter Bándi, Rob van de Loo, Milad Intezar, Daan Geijs, Francesco Ciompi, Bram van Ginneken, Jeroen van der Laak and Geert Litjens*

Dept. of Pathology and Diagnostic Image Analysis Group, Radboud University Medical Center, Nijmegen, The Netherlands

## ABSTRACT

Tissue segmentation is an important pre-requisite for efficient and accurate diagnostics in digital pathology. However, it is well known that whole-slide scanners can fail in detecting all tissue regions, for example due to the tissue type, or due to weak staining because their tissue detection algorithms are not robust enough. In this paper, we introduce two different convolutional neural network architectures for whole slide image segmentation to accurately identify the tissue sections. We also compare the algorithms to a published traditional method. We collected 54 whole slide images with differing stains and tissue types from three laboratories to validate our algorithms. We show that while the two methods do not differ significantly they outperform their traditional counterpart (Jaccard index of 0.937 and 0.929 vs. 0.870, $p < 0.01$).

## 1. INTRODUCTION

Digital pathology is opening new avenues for pathologists. Straightforward archiving, remote diagnostics and application of image analysis to improve efficiency of the diagnostic process are among the most commonly mentioned advantages of digital pathology [1].

Although these advantages sound promising, a digital workflow comes with its own challenges. To generate diagnostic images, whole-slide scanners are used to digitize glass slides containing tissue specimens. Whole-slide scanners try to identify all the areas of tissue on the histopathological slides to decide which areas to scan and to determine the correct focus depth for those areas. However, it is well known that scanners can fail in detecting all tissue regions, for example due to the tissue type (e.g. fatty tissue), or due to weak staining (e.g. in immunohistochemistry). Missed regions can be hugely important for diagnostics, for example when pathologists are looking for cancer metastases in sentinel lymph nodes. Furthermore, accurate tissue segmentation is often an important first step in computerized analysis of digital pathology images.

Unfortunately, there is no way to recover from errors made in tissue detection by slide scanners in later steps of the digital pathology workflow. The easiest solution, to scan every part of the slide completely, is not feasible in clinical practice, as it would increase the scan time and file size beyond reasonable limits. Currently, in many diagnostic settings a technician checks every slide after scanning for quality control. This is a tedious and expensive procedure, were the technician must manually identify the tissue areas in a coarse overview image and subsequently must re-scan those areas. As an alternative, we propose to use automated image analysis algorithms to identify tissue areas.

Some groups have already tried to design methods to improve the tissue detection in scanners. Bug et al. used a method based on global thresholding at the mean value of the Gaussian blurred Laplacian of the grayscale image [2]. The result is subsequently refined via flood filling from identified background points. Hiary et al. built a different algorithm based on k-means clustering using pixel intensity, color and texture features [3].

In recent years, several papers have been published showing the potential of deep learning in digital histopathology [4][5][6]. These results motivated us to assess the value of deep learning in tissue segmentation in digital pathology.

In this paper, we compare the traditional image analysis method from Bug et al., a fully convolutional deep learning approach and a U-net based deep learning approach with respect to tissue segmentation accuracy [7].

## 2. MATERIALS

We used 54 histopathological whole slide images of four different tissue types with hematoxylin and eosin (H&E) and three different immunohistochemical staining with 3,3'-Diaminobenzidine (DAB) chromogen and haemotoxylin counterstaining from three different laboratories (overview in Table 1). This allowed us to cover almost all use-cases and variations which one would encounter in regular clinical practice.

The images were scanned with 3DHistech Pannoramic 250 Flash and Hamamatsu NanoZoomer 2.0 HT C9600-13 whole-slide scanners and stored in vendor specific multi-resolution format. Their size was approximately

100000×220000 pixels with 0.24×0.24 µm and 0.23×0.23 µm pixel spacing respectively.

| Tissue | Staining | Images |
|---|---|---|
| breast | H&E | 8 |
| breast | IHC | 6 |
| lymph node | H&E | 12 |
| lymph node | IHC | 5 |
| rectum | H&E | 4 |
| tongue | H&E | 8 |
| tongue | IHC | 11 |

*Table 1* Whole slide image data set collected from three different labs.

The tissue areas were manually annotated and we have divided the data set randomly into two groups of 27 images for two-fold cross validation. Each training fold was further subdivided into 19 + 8 images for training and validation respectively.

## 3. METHODS

We implemented the Foreground Extraction from Structure Information (FESI) method as a traditional image analysis baseline [2]. Additionally, we trained two different convolutional neural networks for tissue-background segmentation. The first network is a fully convolutional neural network (FCNN) while the second is based on the U-Net network architecture (UCNN) [7].

We quantitatively compared the algorithms using the Jaccard index and we also performed a qualitative assessment to get an understanding of the type of errors the different algorithms make. The Jaccard index compares the tissue mask obtained from the algorithms to the manually annotated ground truth via the following equation:

$$J(A, B) = \frac{|A \cap B|}{|A \cup B|},$$

where A and B are the set of pixels labeled as tissue by two different segmentations.

In the following subsection, we will detail the methodology of the three algorithms.

### 3.1. Foreground Extraction from Structure Information

The FESI algorithm was applied to the 5$^{th}$ resolution level of the multiresolution images where the pixel spacing was 7.68×7.68 µm as it was published [2]. The color images were first converted to grayscale by summing the red, green and blue channels with 0.299, 0.587 and 0.114 weights respectively. The absolute value of the Laplacian was calculated for the grayscale images and blurred by a strong Gaussian filter with 15×15 pixels kernel size and $\sigma = 4$. An initial segmentation was calculated by applying a global threshold at the mean intensity value of the blurred image. Then median blurring with 45×45 pixels kernel size and morphological opening were used for further refining the initial mask. The morphological opening step consisted of 5 erosion followed by 5 dilation steps with a circular kernel with diameter of 7 pixels.

Next, to remove holes from the tissue mask distance transformation was applied on the inverse of the initial segmentation and the point with the maximal value was used as seed point for flood filling the background. The selected point was the farthest away from the already identified tissue regions therefore it reliably belonged to the background.

Finally, the small tissue regions were removed depending on their diameter and distance from large tissue blocks by calculating a distance transformation on the current tissue mask and iterating through the maxima. In each step the pixel with the maximal value of the distance image was taken as seed point. If the value of the seed point was larger than 100 or the seed point was closer than 100 pixels to any previously accepted seed point the point was added to the list of accepted seed points and its containing region was marked as tissue, otherwise the containing region was discarded from the tissue regions and marked as background.

### 3.2. Fully convolutional neural network architecture

Our fully convolutional neural network (FCNN) consisted of 7 convolutional layers with filter sizes 5×5 in the first two convolutional layers, 3×3 on the third and fourth layers, 11×11 on the fifth layer and 1×1 on the last two layers. The number of filters were 16, 32, 64, 64, 1024, 512 and 2 respectively. Max pooling with 2×2 pooling size and stride of 2 was inserted after each of the first three convolutional layers to reduce the memory requirements of the network.

### 3.3. U-Net architecture

The U-Net architecture (UCNN) consisted of 7 steps. In each step, there were two consecutive convolutional layers with 3×3 filter size. The number of filters was 32 at the first step, doubled in each step on the contraction side and halved at each step on the expansion size of the network. Each step except the last on the contraction side contained max pooling with 2x2 pooling size and stride of 2.

We used the weight matrix of the UCNN to prevent empty parts (i.e. not-scanned) in the whole slide images contributing to the error by setting the weight to 0 where the input patch of the network was empty and to 1 elsewhere.

### 3.4 Network training

For regularization, we used batch normalization on all the convolutional layers for both networks. We also inserted two dropout layers after the last 3 contraction steps in UCNN with $p = 0.5$ dropout probability.

We used categorical cross entropy over softmax as loss and added the L2 loss for regularization with $\lambda = 2 * 10^{-6}$ and $\lambda = 5 * 10^{-7}$ weights for the FCNN and UCNN networks respectively.

Both networks were trained with RGB image patches that were randomly sampled from the images during training and validation.

For both networks the image patches were extracted from the 4th layer of the images where the pixel spacing was 3.84×3.84 µm. The patches were 128×128 pixels for the FCNN. Each patch had a single label (tissue or background) based on the central pixel. For whole-slide segmentation the network was applied in a fully convolutional fashion to the image [8]. The inputs of the UCNN network were patches of 892×892 pixels and the outputs of the U-Net network was a 708×708 segmentation mask. The output of U-Net is smaller than the input due application of valid convolutions in the convolutional layers.

To augment the data set, we applied transformations on each of the extracted patches: mirroring the patch on the horizontal or vertical axis; rotating the patch with 90, 180 or 270 degrees clockwise; blurring the patch with a Gaussian filter with a $\sigma$ from the $[0.1, 0.5]$ interval or applying gamma correction with $\gamma$ from the $[0.5, 1.5]$ interval. Each sampled patch was augmented 4 times with randomly selected augmentation method and parameter.

Each epoch contained 100 training and 100 validation iterations for both networks. In each iteration, we extracted 100 + 100 image patches for the FCNN and 4 + 4 for the UCNN for training and validation. The measured epoch accuracy of the networks was the average accuracy of their validation iterations.

The initial learning rate was $l = 5 * 10^{-4}$ in both cases and it was halved if the accuracy did not improve in 10 epochs. The training stopped when the network did not improve its accuracy on the validation set for 50 consecutive epochs. The FCNN stopped after 172 and 165 epochs in the two folds respectively while the UCNN stopped after 196 and 137 epochs.

We used the Adam method for parameter optimization with the He method for initialization in both cases [9][10].

We also implemented a selective sampling method for the UCNN network architecture [11]. All the patches for training in the given epoch were pre-extracted from the slides. For each iteration in the epoch the patches for the training batches were selected randomly from this collection based on a probability that was initialized to 1.0 and updated according to the output of the network. The selection probability update rule was $p_{sel} = 1 - a$, where $a$ was the average classification accuracy given by the network for the patch. The best classified 90% of the patches were replaced with new ones at the end of the epochs.

The trained networks were applied to entire slides on a tile-by-tile basis resulting in a tissue likelihood map. To generate the final binary segmentation mask, we applied Gaussian smoothing with $\sigma = 1.0$ and thresholded the result at a likelihood of 0.5.

## 4. RESULTS

To compare the different methods, we used the Jaccard index (JI). Summary statistics of the Jaccard index across all 54 slides are presented in Table 2 and Figure 1. Some qualitative segmentation results are shown in Figure 3. One can appreciate that both deep learning methods outperform the traditional method on average (JI of 0.929 and 0.937 vs. 0.870). More importantly, they also show less outliers and more stable results (JI standard deviation of 0.059 and 0.063 vs. 0.148).

| Method | Jaccard index mean | Jaccard index standard deviation |
|--------|-------------------|----------------------------------|
| FESI   | 0.870             | 0.148                            |
| FCNN   | 0.937             | 0.063                            |
| UCNN   | 0.929             | 0.059                            |

*Table 2 Summary statistics of Jaccard index for all three methods.*

We also performed a repeated measures ANOVA with a Greenhouse-Geisser correction to assess the statistical significance of the difference in performance. We found that the Jaccard index mean differed statistically significantly between FESI and the deep learning methods ($p < 0.001$). Although the mean Jaccard index of FCNN (0.937) was higher than of UCNN (0.929) no statistically significant difference was found ($p = 0.18$).

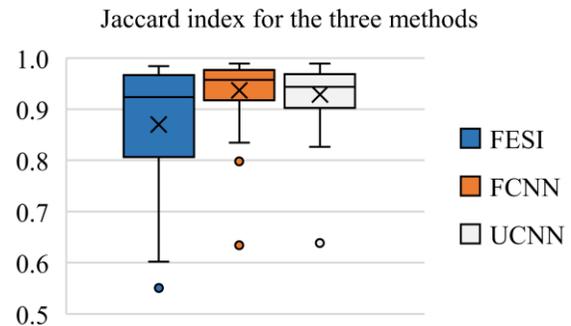

*Figure 1 Visualizing the Jaccard index as a box plot. The X indicates the mean.*

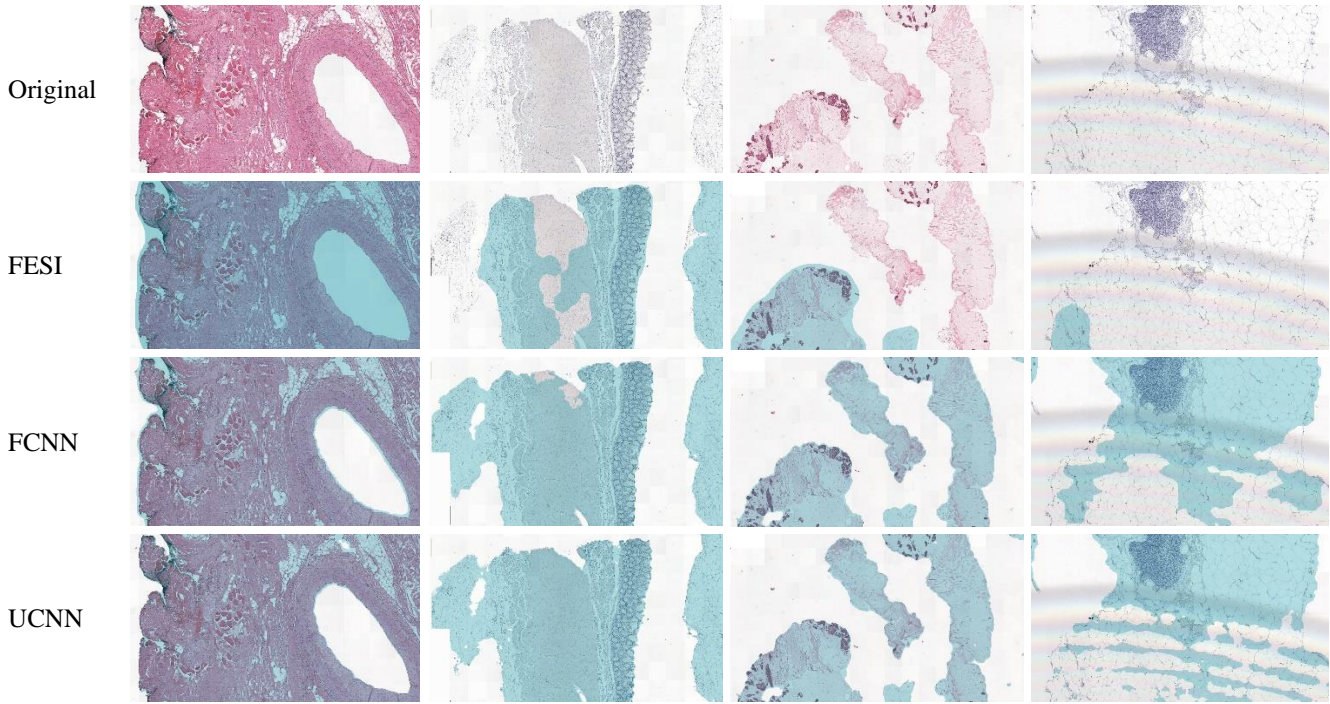

*Figure 2* Qualitative results for the different algorithms. Column 1: zoomed-in tissue section on which all methods performed well. The deep learning algorithms are also able to identify the larger holes in the tissue as background correctly. Column 2: A full whole slide segmentation where the CNNs did much better than the FESI algorithm, here the UCNN gave the best results. Note that the fatty tissue (the web-like structure) on the left side is especially well segmented by the CNNs. Column 3: challenging whole-slide image with weak staining. The FESI algorithm fails to identify all the relevant tissue. Column 4: an example with fatty tissue and a slide preparation artifact where all three algorithms failed to segment the tissue correctly.

## 5. DISCUSSION

The proposed FCNN and UCNN method performed well on the data set; unlike the FESI algorithm the deep learning approaches performed well on different tissue types and stains from all three different laboratories. Although the overall performance of both deep learning algorithms was excellent, there were some difficult tissue areas where all methods made mistakes like fatty tissue areas under air bubbles (last column in Figure 2).

We compared our new methods to the existing FESI method and noticed that the Jaccard index differed significantly from the reported value in the original (0.95 in 34 out of 43 cases). This is probably because in our tests the algorithm was challenged with images from different laboratories and with differences in stains (e.g. we included immunohistochemistry). FESI results could probably be improved by changing parameters on a slide-by-slide basis, but this would essentially make it a semi-automatic method. Furthermore, slide-by-slide optimization is not needed for the deep learning algorithms.

We did not directly compare to the other published tissue segmentation methods by Hiary et al. [3]. They report their results as a localization error in which a pathologist partly determined which errors were relevant (and thus counted), making the approach irreproducible for us. Re-implementing this method was not feasible due to missing algorithmic details.

To further improve our algorithms, we would like to collect and annotate more slides. This would help to identify rare slide processing artifacts like air bubbles in tissue areas. We would also like to incorporate stain normalization methods to make the algorithms even more robust to staining and scanning differences. In our current implementation, we introduce the misclassified patches to the networks repeatedly. However, the spatial distribution of the difficult patches is not taken into account (e.g. the patch next to a difficult patch is probably also difficult). Considering it could further help the networks focus on challenging areas.

The computation time for the FCNN and UCNN was on average 2 and 4 minutes per slide, respectively. Those are larger than the 5 second execution time of FESI but still acceptable.

Concluding, we proved the usability of two different deep learning methodologies in tissue segmentation of whole slide imaging and showed that they can significantly outperform the existing traditional image analysis algorithms.

## 6. REFERENCES

[1] D. R. J. Snead, Y-W. Tsang, A. Meskiri, P. K. Kimani, R. Crossman, N. M. Rajpoot, E. Blessing, K. Chen, K. Gopalakrishnan, P. Matthews, N. Momtahan, S. Read-Jones, S. Sah, E. Simmons, B. Sinha, S. Suortamo, Y. Yeo, H. El Daly, and I. A. Cree. "Validation of digital pathology imaging for primary histopathological diagnosis.", *Histopathology*, vol. 68, pp. 1063–1072, 2016.

[2] D. Bug, F. Feuerhake, and D. Merhof. "Foreground extraction for histopathological whole-slide imaging.", *Bildverarbeitung für die Medizin 2015*, pp. 419–424, 2015.

[3] H. Hiary, R. S. Alomari, and V. Chaudhary. "Segmentation and localisation of whole slide images using unsupervised learning.", *Image Processing, IET*, vol. 7, pp. 464–471, 2013.

[4] G. Litjens, C. I. Sánchez, N. Timofeeva, M. Hermsen, I. Nagtegaal, I. Kovacs, C. Hulsbergen-van de Kaa, P. Bult, B. van Ginneken, and J. van der Laak. "Deep learning as a tool for increased accuracy and efficiency of histopathological diagnosis.", *Nat Sci Rep*, vol. 6: 26286, 2016.

[5] B. Ehteshami Bejnordi, M. Balkenhol, G. Litjens, R. Holland, P. Bult, N. Karssemeijer, and J. van der Laak. "Automated detection of DCIS in whole-slide H&E stained breast histopathology images.", *IEEE Trans Med Imaging*, vol. 35, pp. 2141–2150, 2016.

[6] G. Litjens, B. Ehteshami Bejnordi, N. Timofeeva, G. Swadi, I. Kovacs, C. A. Hulsbergen-van de Kaa, and J. A. W. M. van der Laak. "Automated detection of prostate cancer in digitized whole-slide images of H&E-stained biopsy specimens.", *Medical Imaging*, vol. 9420 of *Proceedings of the SPIE*, p. 94200B, 2015.

[7] O. Ronneberger, P. Fischer, and T. Brox. "U-net: Convolutional networks for biomedical image segmentation.", *Med Image Comput Comput Assist Interv*, vol. 9351 of *Lect Notes Comput Sci*, pp. 234–241, 2015.

[8] J. Long, E. Shelhamer, and T. Darrell. "Fully convolutional networks for semantic segmentation.", *arXiv:14114038*, 2015.

[9] D. Kingma and J. Ba. "ADAM: A method for stochastic optimization.", *arXiv:14126980*, 2015.

[10] K. He, X. Zhang, S. Ren, and J. Sun. "Delving deep into rectifiers: Surpassing human-level performance on imagenet classification.", *Comput Vis Pattern Recognit*, pp. 1026–1034, 2015.

[11] M. J. J. P. van Grinsven, B. van Ginneken, C. B. Hoyng, T. Theelen, and Sánchez C. I. "Fast convolutional neural network training using selective data sampling: Application to hemorrhage detection in color fundus images.", *IEEE Trans Med Imaging*, vol. 35, pp. 1273–1284, 2016.